\documentclass[letterpaper]{article} 
\usepackage{aaai}
\usepackage{times}
\usepackage{helvet}
\usepackage{courier}
\usepackage{theorem}
\usepackage{xspace}
\usepackage{boxedminipage}
\usepackage{graphicx}
\usepackage{amssymb}
\usepackage{amsmath}
\usepackage{amsfonts}
\usepackage{alltt}
\usepackage{array}
\usepackage{latexsym}
\usepackage{color}
\usepackage{url}
\usepackage{soul}
\usepackage{algorithm}
\usepackage{epstopdf}
\usepackage[noend]{algorithmic}
\usepackage{subfigure}

\algsetup{indent=2em}
\newcommand{\fs}[1]{\fontsize{#1}{#1}\selectfont}

\makeatletter
\newcommand{\nbcite}{\def\citeauthoryear##1##2{\def\@thisauthor{##1}%
\ifx \@lastauthor \@thisauthor \relax \else##1 \fi ##2}\@nbcite}
\def\citeS{\@ifnextchar[{\@jbciteS}{\@jbciteS[]}}
\def\@jbciteS[#1]#2{%
\ifthenelse{\equal{#1}{}}{%
\citeauthor{#2}'s (\citeyear{#2})}{%
\citeauthor{#2}'s #1 (\citeyear{#2})}}
\makeatother

\newcommand{\eat}[1]{}

\def\noop{\emph{discard}\xspace}

\def\next{\textbf{next}}

\def\shop2{\textbf{{\fs{8.5}\textsc{SHOP2}}}}

\def\next{\textbf{next}}

\def\tab{\hspace*{.25in}}

\newcounter{annocount}
  {\end{alltt}}

\begin{document}

\title{Knowledge Engineering for Planning-Based Hypothesis Generation}

\author{Shirin Sohrabi ~~~~ Octavian Udrea ~~~~ Anton V. Riabov \\
IBM T.J. Watson Research Center \\ 
PO Box 704, Yorktown Heights, NY 10598, USA\\
\{ssohrab, oudrea, riabov\}@us.ibm.com
}

\nocopyright
\maketitle

\begin{abstract}
In this paper, we address the knowledge engineering problems for hypothesis generation motivated by applications that require timely exploration of hypotheses under unreliable observations. We looked at two applications: malware detection and intensive care delivery. In intensive care, the goal is to generate plausible hypotheses about the condition of the patient from clinical observations and further refine these hypotheses to create a recovery plan for the patient. Similarly, preventing malware spread within a corporate network involves generating hypotheses from network traffic data and selecting preventive actions.
To this end, building on the already established characterization and use of AI planning for similar problems, we propose use of planning for the hypothesis generation problem. However, to deal with uncertainty, incomplete model description and unreliable observations, we need to use a planner capable of generating multiple high-quality plans. To capture the model description we propose a language called LTS++ and a web-based tool that enables the specification of the LTS++ model and a set of observations. We also proposed a 9-step process that helps provide guidance to the domain expert in specifying the LTS++ model. 
The hypotheses are then generated by running a planner on the translated LTS++ model and the provided trace. 
The hypotheses can be visualized and shown to the analyst or can be further investigated automatically.

\end{abstract}

\setlength\floatsep{0.2cm}
\setlength\textfloatsep{0.4cm}
\setlength\belowcaptionskip{0.05cm}
\setlength\theorempreskipamount{6pt plus 1pt minus 0.5pt}
\setlength\theorempostskipamount{6pt plus 1pt minus 0.5pt}


\section{Introduction}

\setlength{\tabcolsep}{1pt} 

\begin{figure*}[ht!]
\begin{center}
\begin{tabular}{ccc}
\includegraphics[width=0.975\columnwidth]{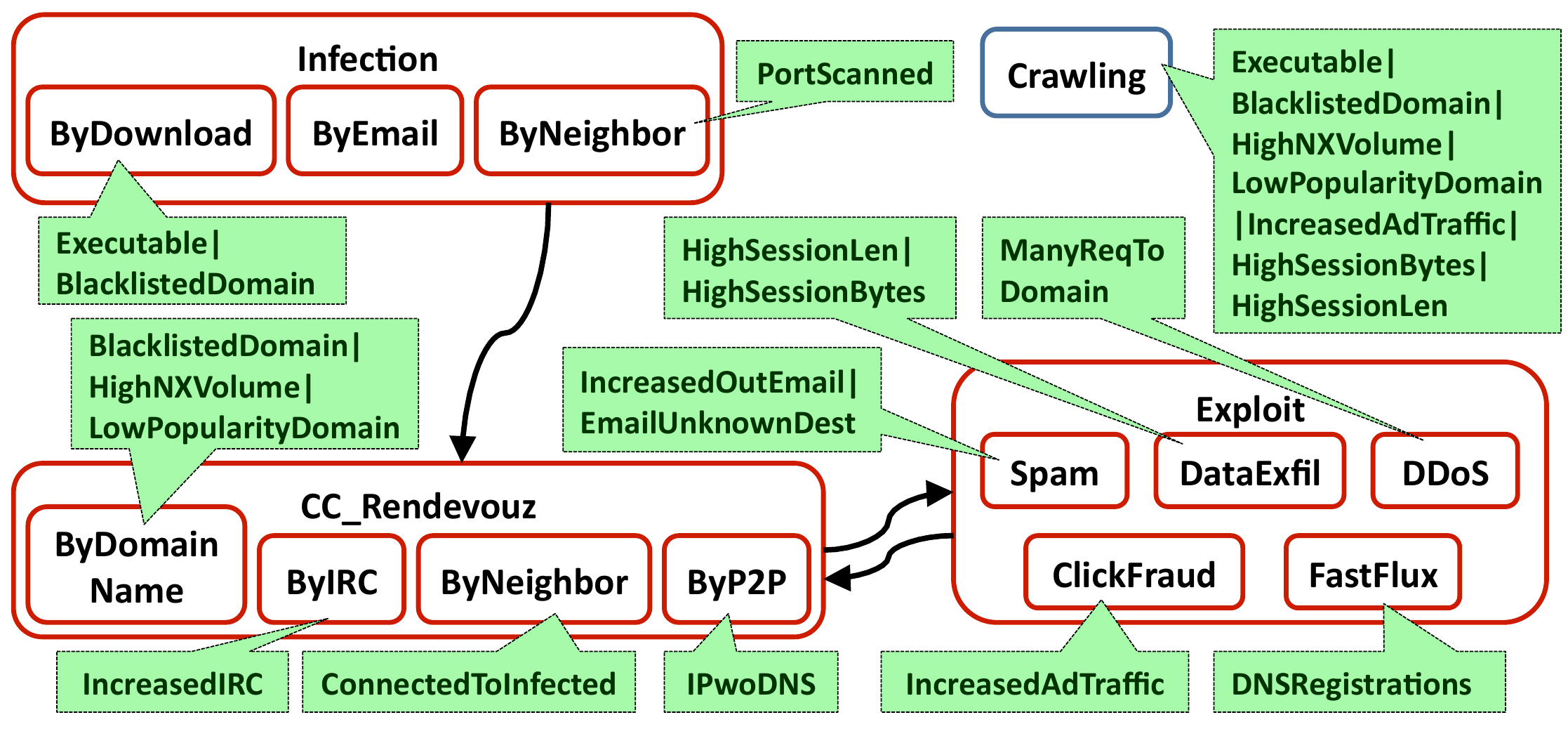} & $\tab$ $\,\,$&
\includegraphics[width=0.955\columnwidth]{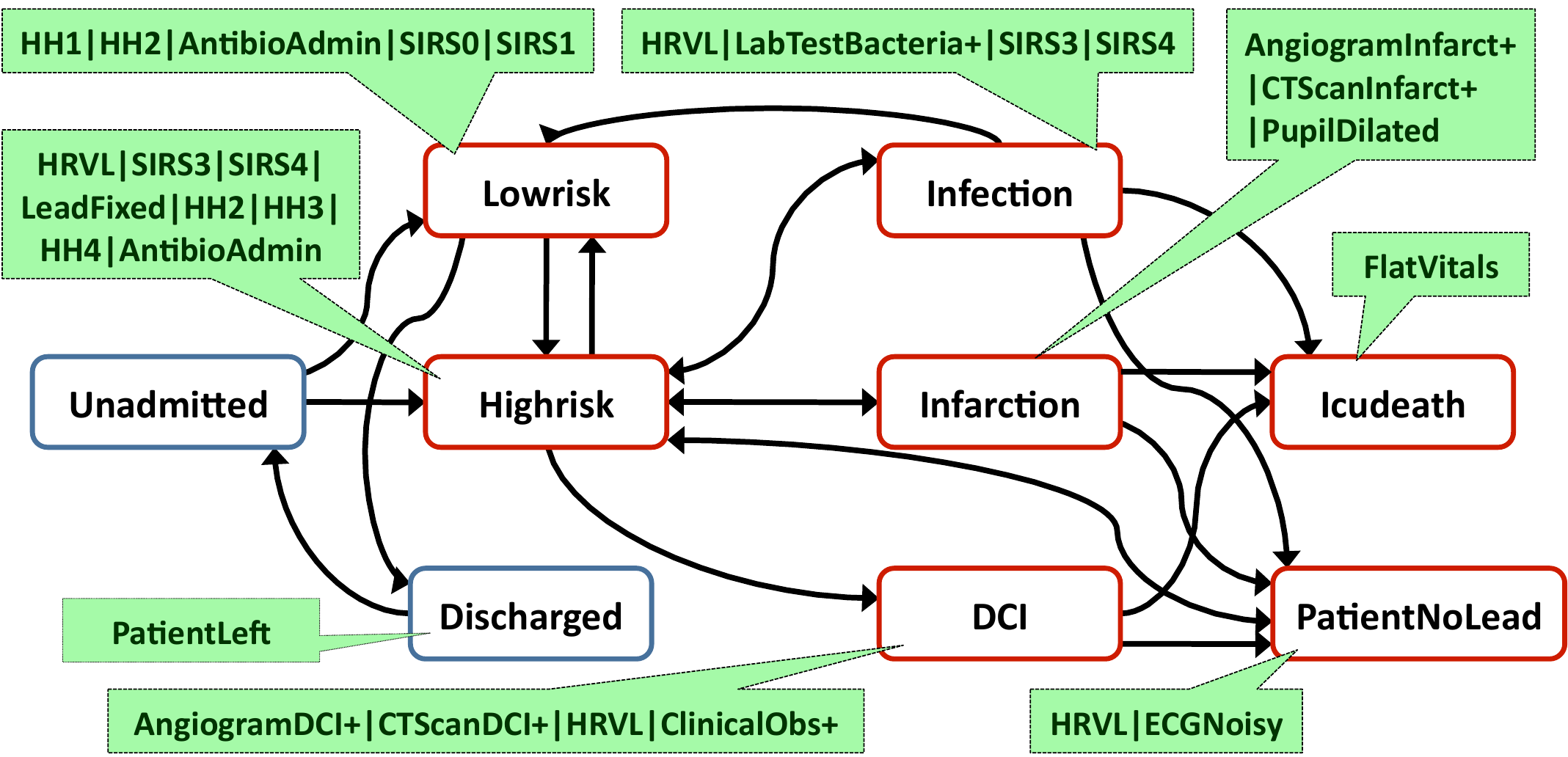} \\
Figure 1 (a): Malware detection &  & Figure 1 (b):  Intensive care \\
\end{tabular}

\label{fig:ex-lts}
\vspace{-0.34cm}
\end{center}
\end{figure*}

Several application scenarios require the construction of hypotheses presenting alternative explanation of a sequence of possibly unreliable observations. 
For example, the evolution of the state of the patient over time in an Intensive Care Unit (ICU) of a hospital can be inferred from a variety of measurements. Similarly, observations from network traffic can indicate possible malware. The hypotheses, represented as a sequence of changes in patient state, aim to present an explanation for these observations, while providing deeper insight into the actual underlying causes for these observations, helping to make decisions about further testing, treatment or other actions. 

Expert judgment is the primary method used for generating hypotheses and evaluating their plausibility. Automated methods have been proposed, to assist the expert, and help improve accuracy and scalability. Notably, model-based diagnosis methods can determine whether observations can be explained by a model (e.g., \cite{Cassandras99,sampath95}).
Recently, several researchers have proposed use of automated planning technology to address several related class of problems including diagnosis (e.g., \cite{sohBiaMcIKR10,Haslum-Two-11}), plan recognition \cite{RamirezG09}, and finding excuses \cite{GobelbeckerKEBN10}. These problems share a common goal of finding a sequence of actions that can explain the set of observations given the model-based description of the system. However, most of the existing literature make an assumption that the observations are all perfectly reliable and should be explainable by the system description, otherwise no solution exists for the given problem. 
But that is not true in general. 
For example, even though observations resulting from the analysis of network data can be unreliable, we would still like to explain as many observations as possible with respect to our model; as a further complication, we cannot assume the model is complete.

In 2011, Sohrabi et al.  established a relationship between generating explanations, a more general form of diagnosis, and a planning problem \cite{sohBiaMcIAAAI2011}. Recently, we extend this work to address unreliable observations and showed how to generate multiple high-quality plans or the plausible hypotheses  \cite{sohUdrRanRiaAAAI13}.
In this paper, we address knowledge engineering problems of capturing the domain knowledge. 

To capture the model description we propose a language called LTS++,  derived from LTS (Labeled Transition System) \cite{lts}
for defining models for hypothesis generation, and associating observation types. LTS++ is less expressive than the general  Planning Domain Definition Language (PDDL) specification of a planning problem \cite{pddl}. However, in our experience, the domain expert finds writing an LTS++ language much simpler than PDDL. 
To further help the domain expert, we also proposed a process that helps provide guidance in specifying the LTS++ model. 
Additionally, we developed a web-based tool that enables the specification of the LTS++ model and a set of observations. 
Our tool features syntax highlighting, error detection, and 
visualization of the state transition graph.
The hypotheses are then generated by running a planner on the translated LTS++ model and the provided observation trace. 
The hypotheses can be visualized and shown to the analyst or can be further investigated automatically.

In the rest of the paper, we will first describe our two application examples in detail. We then describe the architecture of our automated hypothesis exploration problem in which hypothesis generation plays a key role. Then we describe the relationship between planning and hypothesis generation, which facilitates the use of planning technology. We show our initial experimental results in using planning. We then describe our LTS++ language, the creation process, LTS++ IDE, and show several example problems.


\section{Application Description}

In this section we introduce two example applications that illustrate our approach: intensive care delivery and malware detection. A key characteristic of these applications is that the true state of monitored patients, network hosts, or other entities, while essential for timely detection and prevention of critical conditions, is not directly observable. Instead, we must analyze the sequence of available observations to reconstruct the state. To make this possible, our approach relies on a model of the entity consisting of states, transitions between states, and many-to-many correspondence between states and observations. 
In the following sections we will describe how these models can be created by the domain experts and encoded in our LTS++ language.

Figure 1 shows state transition systems of intensive case and malware detection. The rounded rectangles are states. The states are associated with a type, good or bad, and drawn in blue or red respectively. The callouts are observations associated with these states. Note that the observations are obtained by analyzing raw data gathered through sensors. 

In Figure 1 (a), the bad state correspond to malware lifecycle, such as the host becoming infected with malware, the bot's rendezvous with a Command and Control (C\&C) machine (\emph{botmaster}), and a number of exploits -- uses of the bot for malicious activity. Each of the states can be achieved in many ways, depending on the type and capabilities of the malware. For example, the \emph{CC\_Rendezvous} state can be achieved by attempting to contact an Internet domain, or via Internet Relay Chat (IRC) on a dedicated channel.
The good state in Figure 1 (a)  corresponds to a ``normal'' lifecycle of a web crawler compressed into a single state. Note that crawler behavior can also generate a subset of the observations that malware may generate. 
The callouts are the observations associated with states. For example, the observation \emph{HighNXVolume} is an observation associated with the \emph{ByDomainName} state that corresponds to an abnormally high number of \emph{domain does not exist} responses for Domain Name System (DNS) queries; such an observation may indicate that the bot and its botmaster are using a domain name generation algorithm, and the bot is testing generated domain names trying to find its master.

In Figure 1 (b), the bad states correspond to critical states of a patient such as \emph{Infection}, \emph{DCI}, or \emph{Highrisk}. The good states are the non-critical states. Upon admission the patient is either in \emph{Lowrisk} or in \emph{Highrisk}. From a \emph{Highrisk} state, they may get to the \emph{Infection}, \emph{Infarction}, or the \emph{DCI} state. From \emph{Lowrisk} they may get to the \emph{Highrisk} state or be \emph{Discharged} from ICU. The patient enters \emph{Icudeath} from \emph{Infection}, \emph{Infarction}, or \emph{DCI} state. The patient's condition may improve; hence the patient's state may move back to the \emph{Lowrisk} state from for example the \emph{Infection} state. The observations are measured based on the raw data captured by patient monitoring devices (e.g., the patient's blood pressure, heat rate, temperature) as well as other measurements and computations provided by doctors  and nurses. For example, given the patient's heart rate, their blood pressure, and their temperature, which are measured continuously, their SIRS score can be computed, producing an integer between 0 to 4. Similarly, a result of CT Scan, or a lab test will indicate other possible observations about the patient. 

While the complexity of the analysis involved to obtain one observation can vary, it is important to note that observations are by nature \emph{unreliable}:

\emph{The set of observations will be incomplete}. Operational constraints will prevent us running in-depth analysis on \emph{all} of the data all the time. However, all observations are typically time stamped, and hence totally ordered.

\emph{Observations may be ambiguous}. This is depicted in Figure 1, where for instance contacting a blacklisted domain may be evidence of malware activity, or maybe a crawler that reaches such a domain during normal navigation. Similarly, Heart Rate Variability Low (HRVL) may be explained by many states such as \emph{DCI} or \emph{Highrisk} or \emph{Infection}.

\emph{Not all observations will be explainable}. There are several reasons while some observations may remain unexplained: (i) observations are (sometimes weak) indicators of a behavior, rather than authoritative measurements; (ii) the model description is by necessity incomplete, unless we are able to design a perfect model; (iii) in the case of malware detection, malware could try to confuse detectors by either hiding in normal traffic patterns or originating extra traffic.

For Figure 1 (a) one can consider the following two observations for a host: ($o_1$) a download from a blacklisted domain and ($o_2$) an increase in traffic with ad servers. Note that according to Figure 1 (a), this sequence could be explained by two hypotheses: (a) a crawler or (b) infection by downloading from a blacklisted domain, a C\&C rendezvous which we were unable to observe, and an exploit involving click fraud. In such a setting, it is normal to believe (a) is more plausible than (b) since we have no evidence of a C\&C rendezvous taking place. However, take the sequence ($o_1$) followed by ($o_3$) an increase in IRC traffic followed by ($o_2$). In this case, it is reasonable to believe that the presence of malware -- as indicated by the C\&C rendezvous on IRC -- is more likely than crawling, since crawlers do not use IRC. The crawling hypothesis cannot be completely discarded since it may well be that a crawler program is running in background, while a human user is using IRC to chat.

Consider the following observation sequence for the model in Figure 1 (b): \emph{HH3}, \emph{HRVL}. This denotes a patient with a Hunt and Hess (a grading system used to classify the severity of subarachnoid hemorrhage) score of 3, followed by \emph{HRVL}. Since \emph{HRVL} is an ambiguous observation -- i.e., can be indicative of multiple states --, equally plausible hypotheses may be:  \\
$\tab$ \emph{Unadmitted} $\rightarrow$ \emph{Highrisk} or \\
$\tab$ \emph{Unadmitted} $\rightarrow$ \emph{Highrisk} $ \rightarrow$ \emph{PatientNoLead} or \\
$\tab$ \emph{Unadmitted} $\rightarrow$ \emph{Highrisk} $ \rightarrow$ \emph{Infarction} or \\
$\tab$ \emph{Unadmitted} $\rightarrow$ \emph{Highrisk} $ \rightarrow$ \emph{DCI}.

Note, although the current state of the patient is unknown, the generated hypotheses indicate that it is one of \emph{Highrisk}, \emph{PatientNoLead}, \emph{Infarction} or \emph{DCI}.

Given a sequence of observations and the model, the hypothesis generation task infers a number of plausible hypotheses about the evolution of the entity. Practically, we have to analyze multiple hypotheses about an entity because the state transition model may be incomplete or the observations may be unreliable. The result of our automated technique can then be presented to a network administrator (or to a doctor) or to an automated system for further investigation and testing. Next, we will describe briefly all the necessary components for hypothesis exploration.

\subsection{Architecture}

Our work on automated exploration of hypotheses focuses on the Hypothesis Generation, which is part of a larger automated data analysis system that includes sensors, actuators, multiple analytic platforms and a \emph{Tactical Planner}. 
Tactical Planner, 
for  the purpose of this paper, 
should be viewed as a component responsible for execution of certain strategic actions and it can be implemented using, for example, a classical planner to compose analytics~\cite{BouilletMMA09}.
A high-level overview of the complete system architecture is shown in Figure~\ref{fig:architecture}. All components 
of the architecture, 
with the exception of 
application-specific analytics, sensors, and actuators, are designed to be reused without modification in a variety of application domains.

The system receives input from \emph{Sensors}, and \emph{Analytics} translate sensor data to observations. The \emph{Hypothesis Generator} interprets the observations 
received from analytics, 
and generates hypotheses about the state of \emph {Entities} in the \emph{World}. Depending on application domain, the entities may correspond to patients in a hospital, or to computers connected to a corporate network, or other 
kinds of 
objects of interest. 
The \emph{Strategic Planner} evaluates these hypotheses and initiates preventive or testing actions in response. Some of the testing actions can be implemented as additional processing of data, 
setting new goals for the Tactical Planner, which composes and deploys analytics across multiple \emph{Analytic Platforms}. A Hadoop cluster, for example, can be used as an analytic platform for offline analysis of historical data accumulated in one or more \emph{Data Stores}. Alternatively, a Stream Computing cluster can be used for fast online analysis of new data received from the sensors. 

Preventive actions, as well as some of the testing actions, are dispatched to \emph{Actuators}. There is no expectation that every actuation request will succeed, or 
always happen instantaneously. 
Actuation in a hospital setting can involve dispatching alerts to doctors, or lab test recommendations.

\setcounter{figure}{1} 

\begin{figure}[t!]
\begin{center}
\includegraphics[width=0.72\columnwidth]{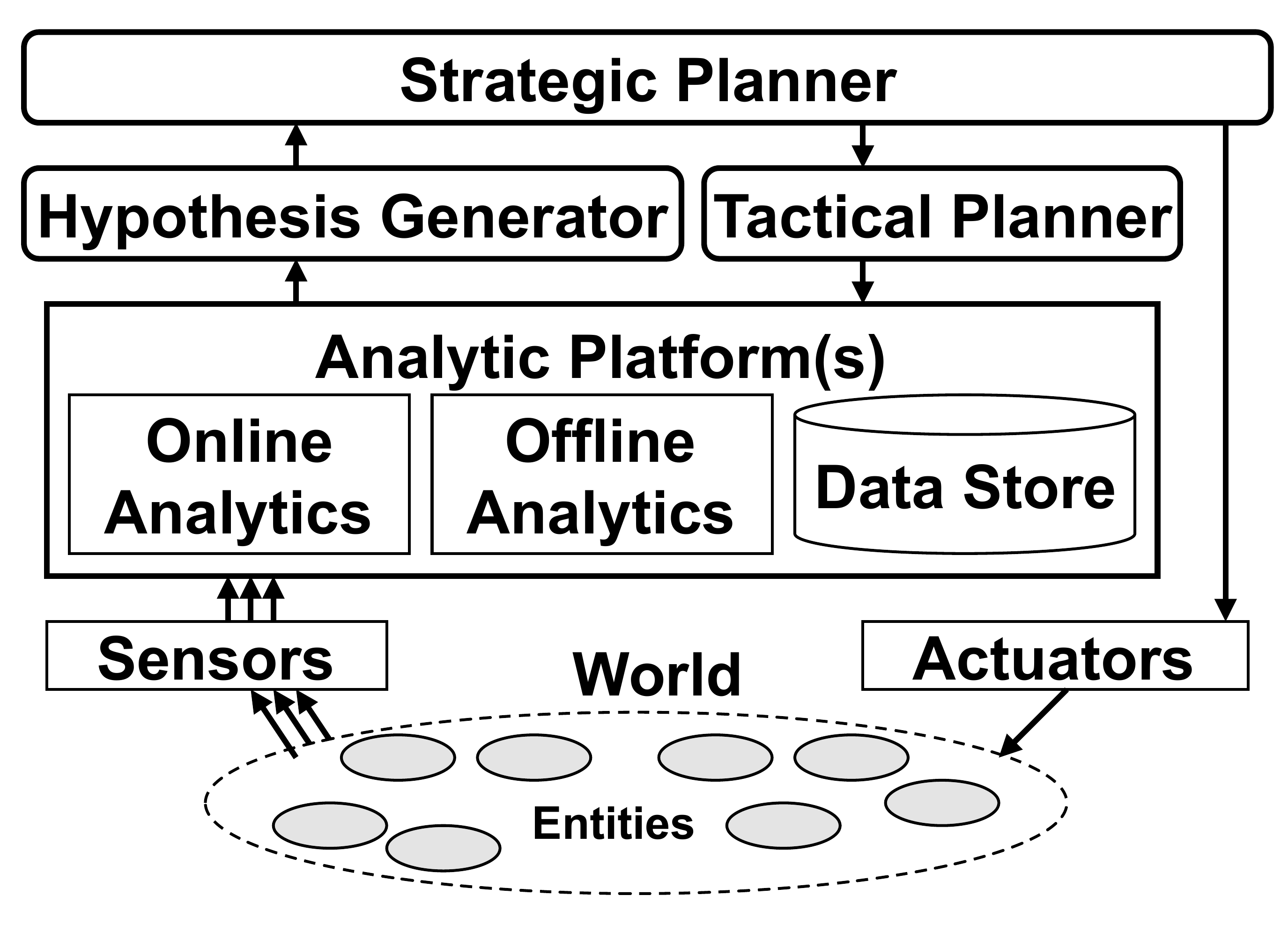}  
\end{center}
\vspace{-0.58cm}
\caption{System architecture}
\label{fig:architecture}
\end{figure}

\eat{We will not describe the implementation of the Tactical Planner in this paper, focusing our attention on the higher-level decisions made by the Strategic Planner. For the purposes of this paper, Tactical Planner should be viewed as a component responsible for execution of certain strategic actions. We note that the Tactical Planner can be implemented using, for example, a classical planner to compose analytics~\cite{BouilletMMA09}.}


\newcommand{\Sys}{\ensuremath{\Sigma}\xspace}
\def\next{\raisebox{1pt}{\begin{footnotesize}\ensuremath{\bigcirc}\end{footnotesize}}}
\newcommand{\eventually}{\begin{large}\ensuremath{\lozenge}\end{large}}
\newcommand{\prefeq}{\ensuremath{\preceq}\xspace}
\newcommand{\pref}{\ensuremath{\prec}\xspace}

\label{planning}

\renewcommand{\arraystretch}{0.95}
\setlength{\tabcolsep}{5pt}

\begin{table*}[t]\footnotesize
{\fs{8.5}
\begin{center}
\begin{tabular}{|c|c|c!{\vrule width 1.2pt}c|c!{\vrule width 1.2pt}c|c!{\vrule width 1.2pt}c|c|}
 \cline{2-9}
 \multicolumn{1}{c|}{}  & \multicolumn{2}{c!{\vrule width 1.2pt}}{\raisebox{-0.4mm}[0mm]{Hand-crafted}} & \multicolumn{2}{c!{\vrule width 1.2pt}}{\raisebox{-0.4mm}[0mm]{10 states}} & \multicolumn{2}{c!{\vrule width 1.2pt}}{\raisebox{-0.4mm}[0mm]{50 states}} & \multicolumn{2}{c|}{\raisebox{-0.4mm}[0mm]{100 states}}\\
 
 \hline
 \rule[1mm]{0mm}{1.5mm} 
Observations & \% Solved & Time & \% Solved & Time & \% Solved & Time & \% Solved & Time \\
\hline
\rule[1mm]{0mm}{1.5mm} 
5	&100\%	&2.49	&70\%	&0.98	&80\%	&5.61	&30\%	&14.21\\
10	&100\%	&2.83	&90\%	&2.04	&50\%	&25.09	&30\%	&52.63\\
20	&90\%	&12.31	&70\%	&24.46	&-		&-		&-		&-\\
40	&70\%	&3.92	&40\%	&81.11  &-		&-		&-		&-\\
60	&60\%	&6.19	&-		&-		&-		&-		&-		&-\\
80	&50\%	&8.19	&-		&-		&-		&-		&-		&-\\
100	&60\%	&11.73	&10\%	&10.87	&-		&-		&-		&-\\
120	&70\%	&20.35	&20\%	&15.66	&-		&-		&-		&-\\
\hline

\end{tabular}
\end{center}
}
\vspace{-0.32cm}
\caption{The percentage of problems where the ground truth was generated, and the average time spent for LAMA.}
\vspace{-0.1cm}
\label{tab:performance}
\end{table*}

\section{Hypothesis Generation via Planning}

In this section, we define the hypothesis generation problem and describe its relationship
to planning. We also provide experimental evaluation that supports the premise of using planning for generating multiple plausible hypotheses. 
In the next section,
we will describe how the planning model can be captured using the LTS++ language which we translate to a planning problem. Our tool, LTS++ hypothesis generator, then uses a planning to compute plausible hypothesis and present them to the user.

Following our recent work \cite{sohUdrRanRiaAAAI13}, a \emph{dynamical system} is defined as $\Sys=(F,A,I)$, where $F$ is a finite set of fluent symbols, $A$ is a set of actions with preconditions and effects that describes actions that account for the possible transitions of the state of the entity (e.g., patient or host)  as well as the \noop action that addresses unreliable observations by allowing observations to be unexplained, and $I$ is a clause over $F$ that defines the initial state. The instances of the \noop action add transitions to the system that account for leaving an observation unexplained. The added transitions ensure that we took all observations into account, but an instance of the  \noop action for a particular observation $o$ indicates that   $o$ is not explained. 
Actions can be over both ``good'' and ``bad'' behaviors. This maps to ``good'' and ``bad'' states of the entity, different from a system state (i.e., set of fluents over $F$).

An observation formula $\varphi$ is a sequence of fluents in $F$ we refer to as \textit{trace}. 
Given a trace $\varphi$, and the system description $\Sys$, 
 a hypothesis $\alpha$ is a sequence of actions in $A$ such that $\alpha$ satisfies $\varphi$ in the system $\Sys$.  
 We also define a notion of plausibility of a hypothesis. 
 Given a set of observations, there are many possible hypotheses, but some could be stated as more plausible than others. 
For example, since observations are not reliable, the hypothesis $\alpha$ can explain a subset of observations by including instances of the  \noop action. However, we can indicate that a hypothesis that includes the minimum number of \noop actions is more plausible.
In addition, observations can be ambiguous: they can be explained by instances of ``good'' actions as well as ``bad'' actions. Similar to the diagnosis problem, a more plausible hypothesis ideally has the minimum number of ``bad'' or ``faulty'' actions.  
More formally, given a system $\Sys$ and two hypotheses $\alpha$ and $\alpha'$ we assume that we can have a reflexive and transitive plausibility relation $\prefeq$, where $\alpha \prefeq \alpha'$ indicates that $\alpha$ is at least as plausible as $\alpha'$.

The \textbf{hypothesis generation problem} is then defined as $P=(F,A',I,\varphi)$ where $A'$ is the set $A$ with the addition of positive action costs that accounts for the plausibility relation $\prefeq$. A hypothesis is a plan for $P,$ and the most plausible hypothesis is the minimum cost plan.  That is, if  $\alpha$ and $\alpha'$ are two hypotheses, where $\alpha$ is more plausible than $\alpha'$, then $cost(\alpha) < cost(\alpha')$. Therefore, the most plausible hypothesis is the minimum cost plan.  

While some class of plausibility relation can be expressed as Planning Domain Definition Language (PDDL3) \cite{pddl3-aij09} preferences, cost-based planners are (currently) more advanced than PDDL3-based planners, and so the technique proposed by Keyder and Geffner \citeyear{KeyderG10} can be used to compile preferences into costs, enabling the use of cost-based planners instead.

\subsection{Computing Plausible Hypotheses}

To address uncertainty, the unreliability of observations and incomplete model description, we must generate multiple high-quality (or low-cost) plans that correspond to a set of plausible hypothesis. 
To this end, we adapt our implementation of hypothesis generation from \cite{sohUdrRanRiaAAAI13}. 
We encode the plausibility notion as actions costs. In particular, we assign a high cost to the \noop  action in order to encourage explaining more observations. In addition, we assign a higher cost to all instances of the actions that represent ``bad'' behaviors than those that represent ``good'' behaviors. Furthermore, shorter/simpler plans are assumed to be more plausible. To address observations, we similarly compile them away in our encoding following a technique proposed in \cite{Haslum-Two-11}. 

The planning problem is described in PDDL. We used one fixed PDDL encoding of the domain, but varied the problem for each problem description, which we generate automatically in our experiments. We also developed a replanning process around LAMA \cite{RichterW10} to generate  multiple high-quality (or low-cost) plans that correspond to a set of plausible hypothesis. The replanning process works in such a way that after each round, the planning problem is updated to disallow finding the same set of plans in future runs of LAMA. This process continues until a time limit is reached and then all found plans are sorted by cost and shown to the user by our tool. 

\eat{
\subsection{Sample Planning Problem}

The planning problems are described in Planning Domain Definition Language (PDDL) \cite{pddl}. One fixed PDDL domain including a total of 6 actions was used in all experiments. 
Actions \emph{explain-observation} and \emph{discard-observation} are used to advance to the next observation in the sequence, and actions \emph{state-change} and \emph{allow-unobserved} change the state of the lifecycle. Two additional actions, \emph{enter-state-good} and \emph{enter-state-bad}, are used to associate different costs for good and bad explanations. 
In our implementation, the good states have lower cost than the bad states: we assume the observed behavior is not malicious until it can only be explained as malicious, and we compute the plausibility of hypotheses accordingly.
The state transitions of malware lifecycle and the observations are encoded in the problem description. This encoding allowed us to automatically generate multiple problem sets that include different number of observations as well as different types of malware lifecycle. 
}

\subsection{Experimental Evaluation}

The experiments we describe in this section help evaluate the response time and the accuracy of our approach. In particular, these experiments show promise of our approach in terms of using planning. This experiments were reported in \cite{sohUdrRanRiaAAAI13}. 
We evaluated performance by using both a hand-crafted description of the malware detection problem and a set of automatically generated state transition systems with 60\% bad and 40\% good states.

To evaluate performance, we introduce the notion of \emph{ground truth}. In all experiments, the problem instances are generated by constructing a ground truth trace by traversing the lifecycle graph (similar to Figure 1 (a)) in a random walk,
adding with small probability, missing and inconsistent observations. 
 We then measure performance by comparing the generated hypotheses with the ground truth, and consider a problem \emph{solved} for our purposes if the ground truth appears among the generated hypotheses.

For each size of the problem, we have generated 10~problem instances, 
and the measurements we present are averages. The measurements were done on a dual-core 3~GHz Intel Xeon processor and 8~GB memory, running 64-bit RedHat Linux. We used a 300 seconds time limit.

Table~\ref{tab:performance} summarizes the result. 
The rows and the columns indicate the problem size, measured by the number of observations and the number of states. The hand-crafted column, is the example shown in Figure 1 (a), which has 18~states. The generated problems consisted of~10, 50 and 100 states. The \emph{\% Solved} column shows the percentage of problems where the ground truth was among the generated plans. The \emph{Time} column shows the average time it took from the beginning of iterations to find the ground truth solution for the solved problems.
The dash entries indicate that the ground truth was not found within the time limit.

The results show that planning can be used successfully to generate hypotheses for malware detection, even in the presence of unreliable observations, especially for smaller size problems. The correct hypothesis was generated in most experiments with up to 10 observations. However, in some of the larger instances LAMA could not find any plans.  Moreover, in the smaller size problems, more replanning rounds is done within the time limit and hence more distinct plans are generated which increases the chance of finding the ground truth.  The results for the hand-crafted malware example also suggest that the problems arising in practice may be easier than randomly generated ones, which had more state transitions and higher branching factor.

We believe that LAMA would have had a better chance of detecting the ground truth trace if instead of finding a set of high-quality plans it could have generated the top $k$ plans, where $k$ could be determined based on a particular scenario. In future work, we plan to evaluate our approach using a planner capable of finding top $k$ plans. Nevertheless, the experiments support our findings, namely, that the use of planning is promising.

\section{LTS++ Model}

To help new users, we have built a web-based tool for generating hypotheses and developing state transition models, which we use in our experiments and applications. In particular, we have designed a language called LTS++, derived from LTS (Labeled Transition System) \cite{lts}, for defining models for hypothesis generation, and associating observation types with states. In this section, we describe a process that the user or the domain expert might undergo in order to define an LTS++ model. We will also describe the LTS++ IDE and the LTS++ syntax.

\begin{figure}[t!]
\begin{center}
\includegraphics[width=1\columnwidth]{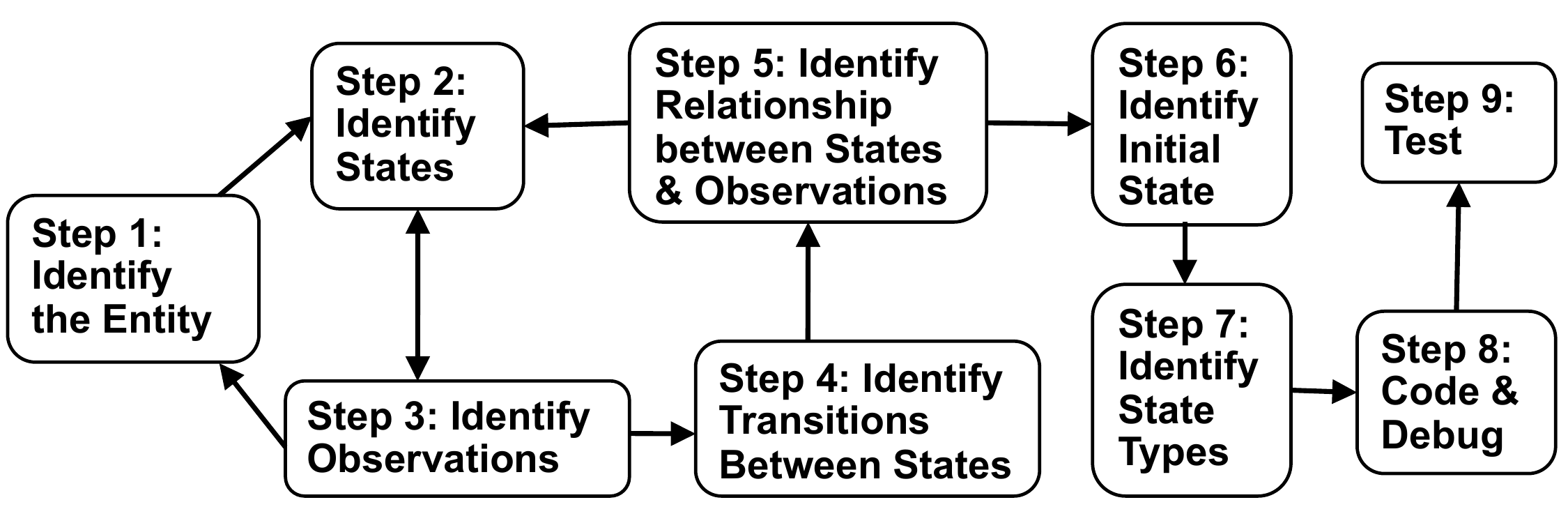}  
\end{center}
\vspace{-0.5cm}
\caption{Process for LTS++ model creation}
\label{fig:process}
\end{figure}

\begin{figure}[t!]
\begin{center}
\includegraphics[width=1\columnwidth]{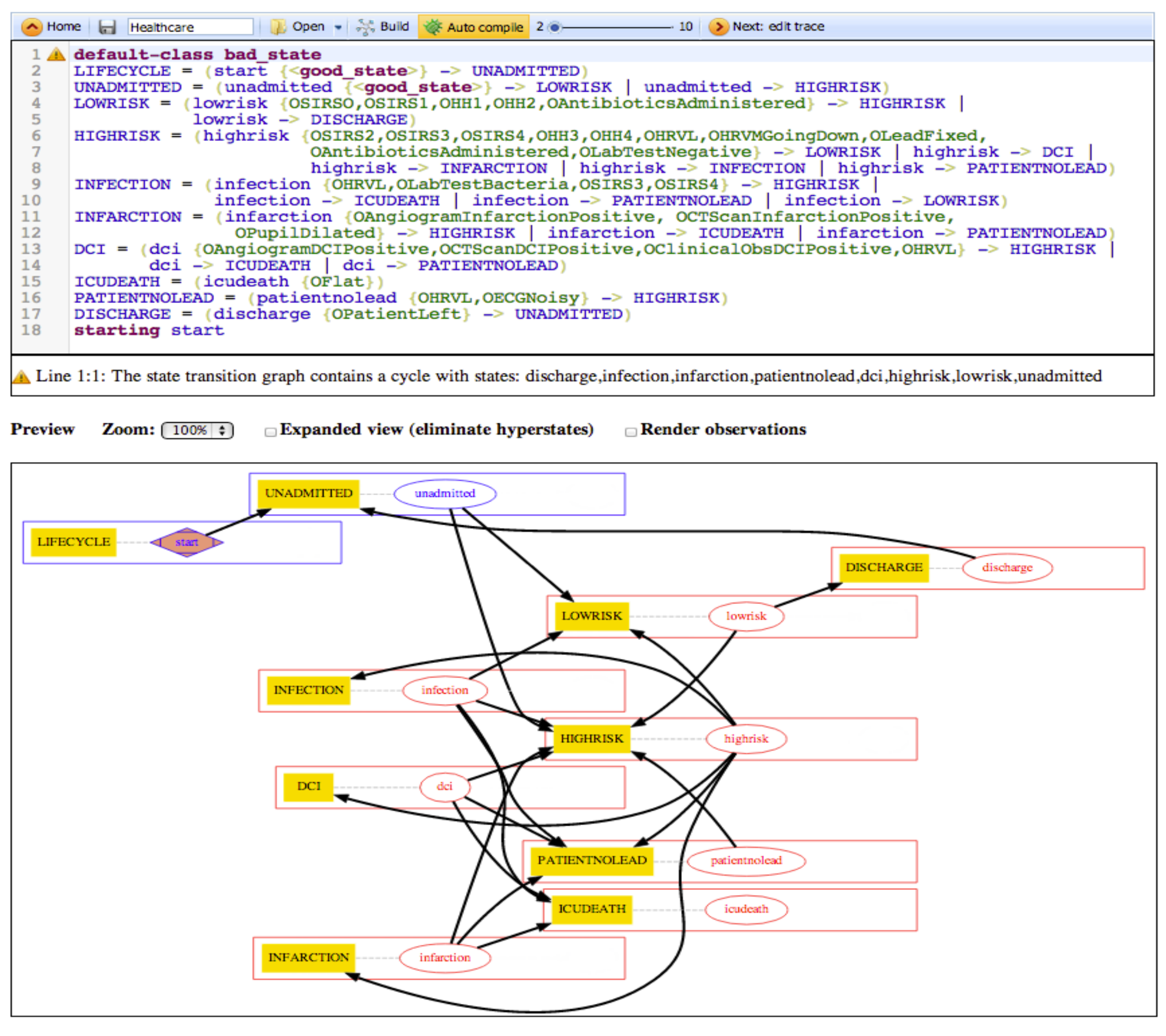}  
\end{center}
\vspace{-0.5cm}
\caption{LTS++ IDE}
\label{fig:ide}
\end{figure}

\begin{figure*}[t!]
\begin{center}
{%
\setlength{\fboxsep}{0pt}%
\fbox{
\includegraphics[width=1.935\columnwidth]{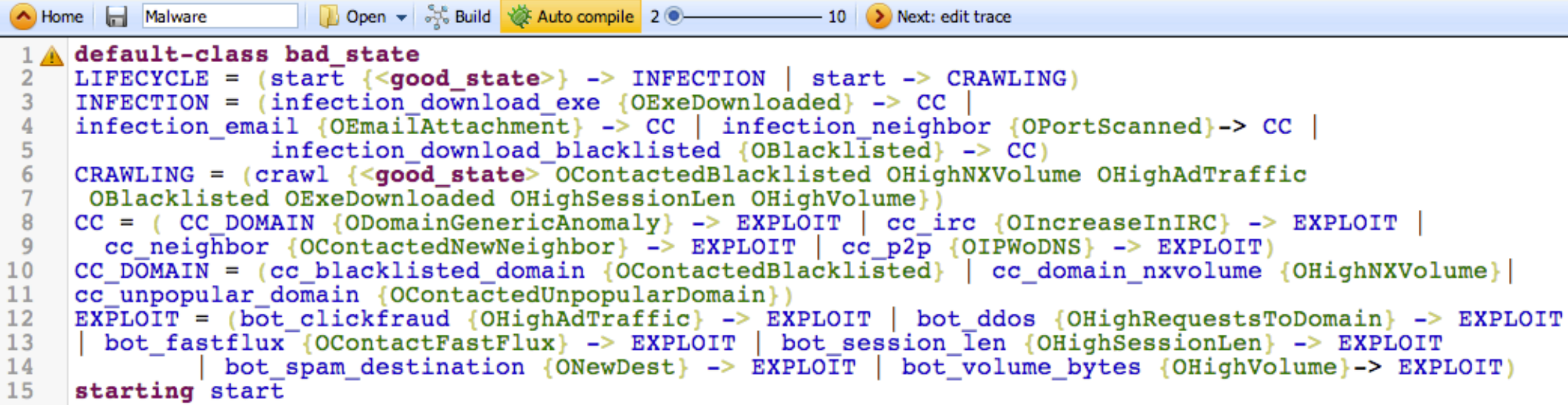}}  
}
\end{center}
\vspace{-0.45cm}
\caption{LTS++ model for malware detection}
\label{fig:cybermodel}
\end{figure*}

\begin{figure*}[t!]
\begin{center}
{%
\setlength{\fboxsep}{0pt}
\fbox{\includegraphics[width=1.95\columnwidth]{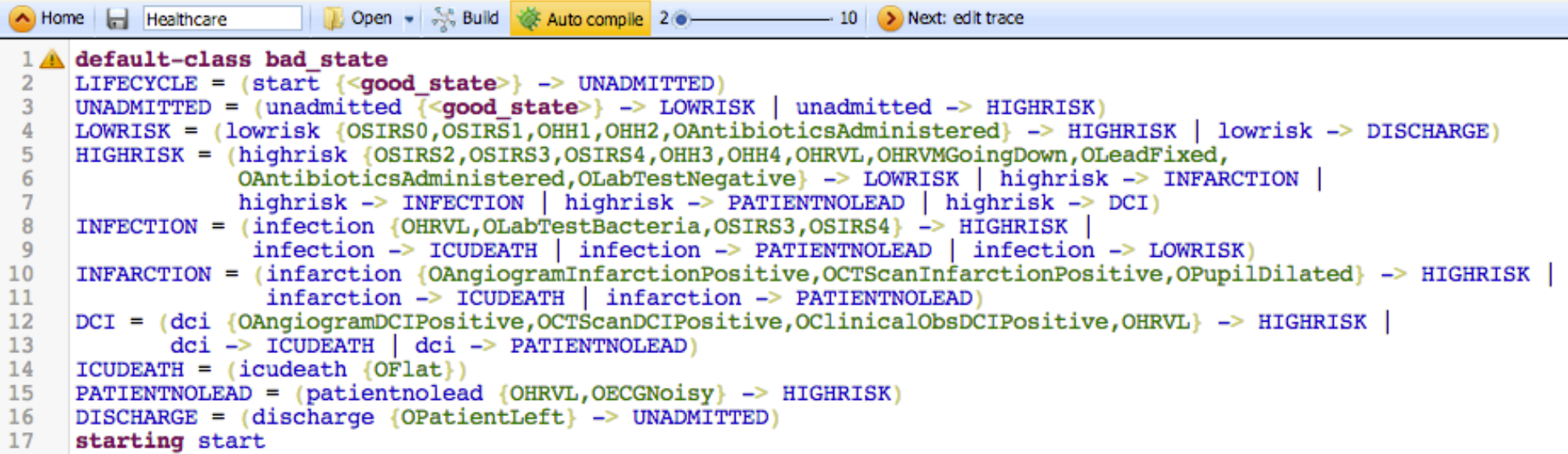}}  
}
\end{center}
\vspace{-0.45cm}
\caption{LTS++ model for the intensive care}
\label{fig:healthcaremodel}
\end{figure*}

\subsection{Steps in Creating an LTS++ Model}

Figure \ref{fig:process} shows a 9-step creation process for an LTS++ model. The arrows are intended to indicate the most typical transitions between steps: transitions that are not shown are not prohibited. This process is meant to help provide guidance to the new users in developing an LTS++ model. In this section, we will go over the first 7 steps.

In step 1, the user needs to identify the entity. This may depend on the objective of the hypothesis generator, the available data, and the available actions. For example, in the malware detection problem, the entity is the host, while 
in the intensive care delivery problem the entity is the patient.
In step 2, the domain expert identifies the states of the entity. As we saw in the application section, the states of patient for example could be for example \emph{DCI}, \emph{Infection}, and \emph{Highrisk}. 
Since the state transition model is manually specified and contains a fixed set of observation types, while potentially trying to model an open world with an unlimited number of possible states and observations, the model can be incomplete at any time, and may not support precise explanations for all observation sequences. To address this on the modeling side, and provide feedback to model designers about states that may need to be added, we have introduced a hierarchical decomposition of states. In some configurations, the algorithm allows designating a subset of the state transition system as a \emph{hyperstate}. In this case, if a transition through one or several states of the hyperstate is required, but no specific observation is associated with the transition, the hyperstate itself is included as part of the hypothesis, indicating that the model may have a missing state within the hyperstate, and that state in turn may need a new observation type associated with it. In the malware detection problem, the \emph{Infection}, \emph{Exploit}, \emph{CC\_rendezvous} are the hyperstates. 

The user needs to identify a set of observations for the particular problem; this is done in step 3. 
The available data, the entity, and the identified states may help define and restrict the space of observations. In step 4, the domain expert has to find out all possible transitions between states. This may be a tedious task, depending on the number of states. However, one can use hyperstates to help manage these transitions. Any transition of the hyper states is carried out to its substates. 
In step 5, the user has to associate observations to states. This associations is shown in Figure 1 using the green callouts. 
In step 6, one can optionally designate a state as the starting state. The domain expert can also 
create a separate starting state that indicates a one of notation by transitioning to multiple states. 
For example, in the malware detection problem, the starting state ``start'' indicates a ``one of'' notation as it transitions to both \emph{Infection} and \emph{Crawling}. 

In step 7, the user can specify state types which indicate that some states are more plausible than the others. State types are related to the ``good'' vs. ``bad'' behaviors and they influence the ranking between hypotheses. For example, the hypothesis that the host is crawling is more plausible than it being infected, given the same trace, which can be explained by both hypotheses.

\setlength{\tabcolsep}{1pt} 
\begin{figure*}[ht!]
{\small
\begin{center}
\begin{tabular}{|c|c|}
\hline
\raisebox{0.2cm}{\includegraphics[width=0.9\columnwidth]{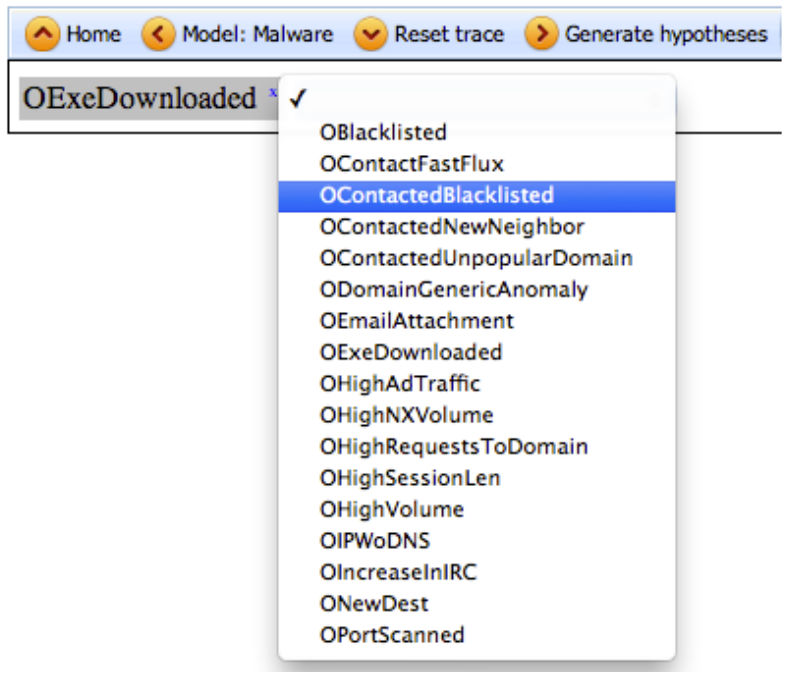}}  &
\raisebox{0.22cm}{\includegraphics[width=0.9\columnwidth]{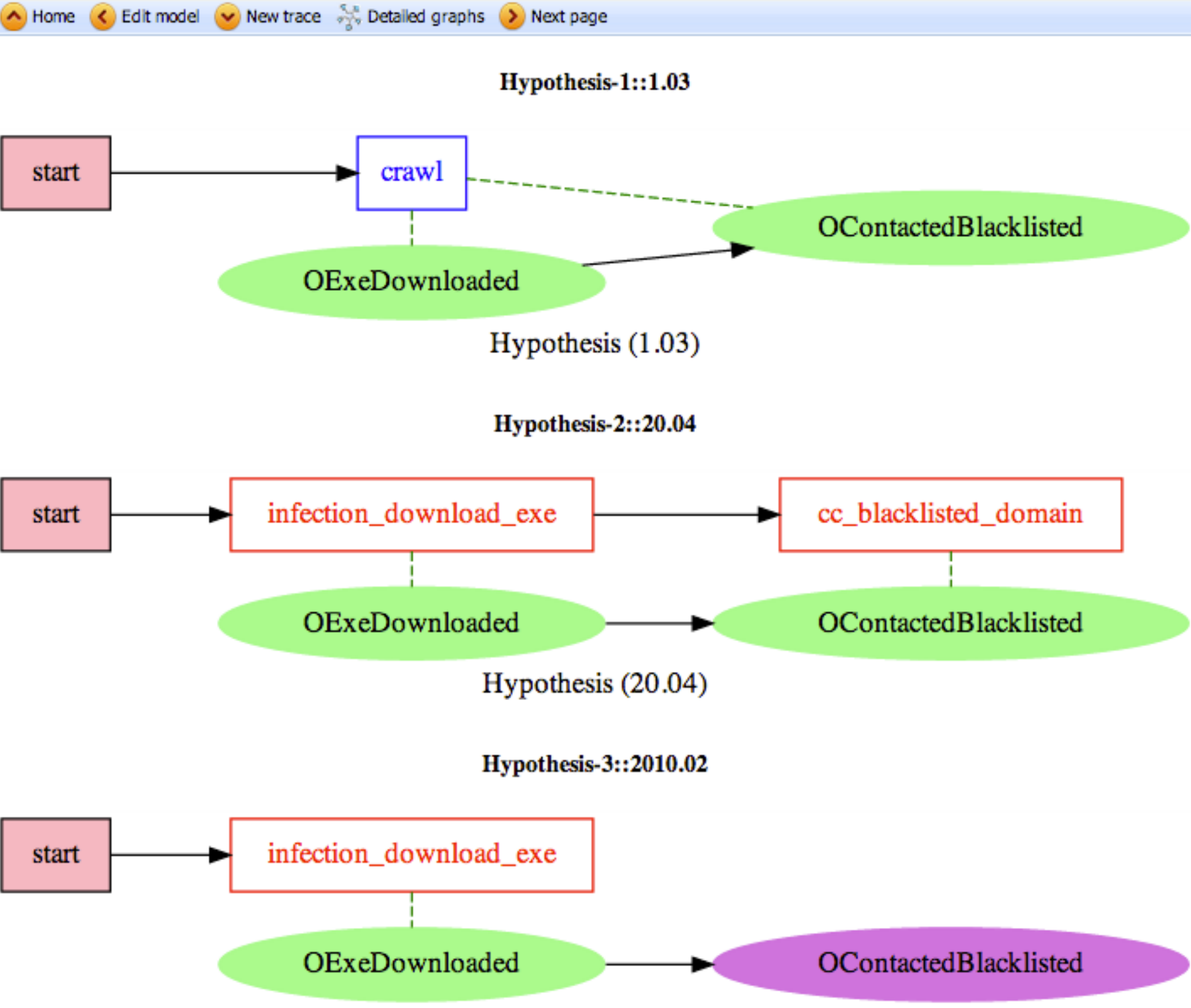}} \\
Figure 7 (a): Entering a trace & Figure 7 (b): Malware example 1 \\
\hline
\end{tabular}

\label{fig:cyberintro}

\vspace{-0.49cm}
\end{center}
}
\end{figure*}

\setlength{\tabcolsep}{1pt} 
\begin{figure*}[ht!]
{\small
\begin{center}
\begin{tabular}{|c|c|}
\hline

\raisebox{+0.0cm}{\includegraphics[width=1.15\columnwidth]{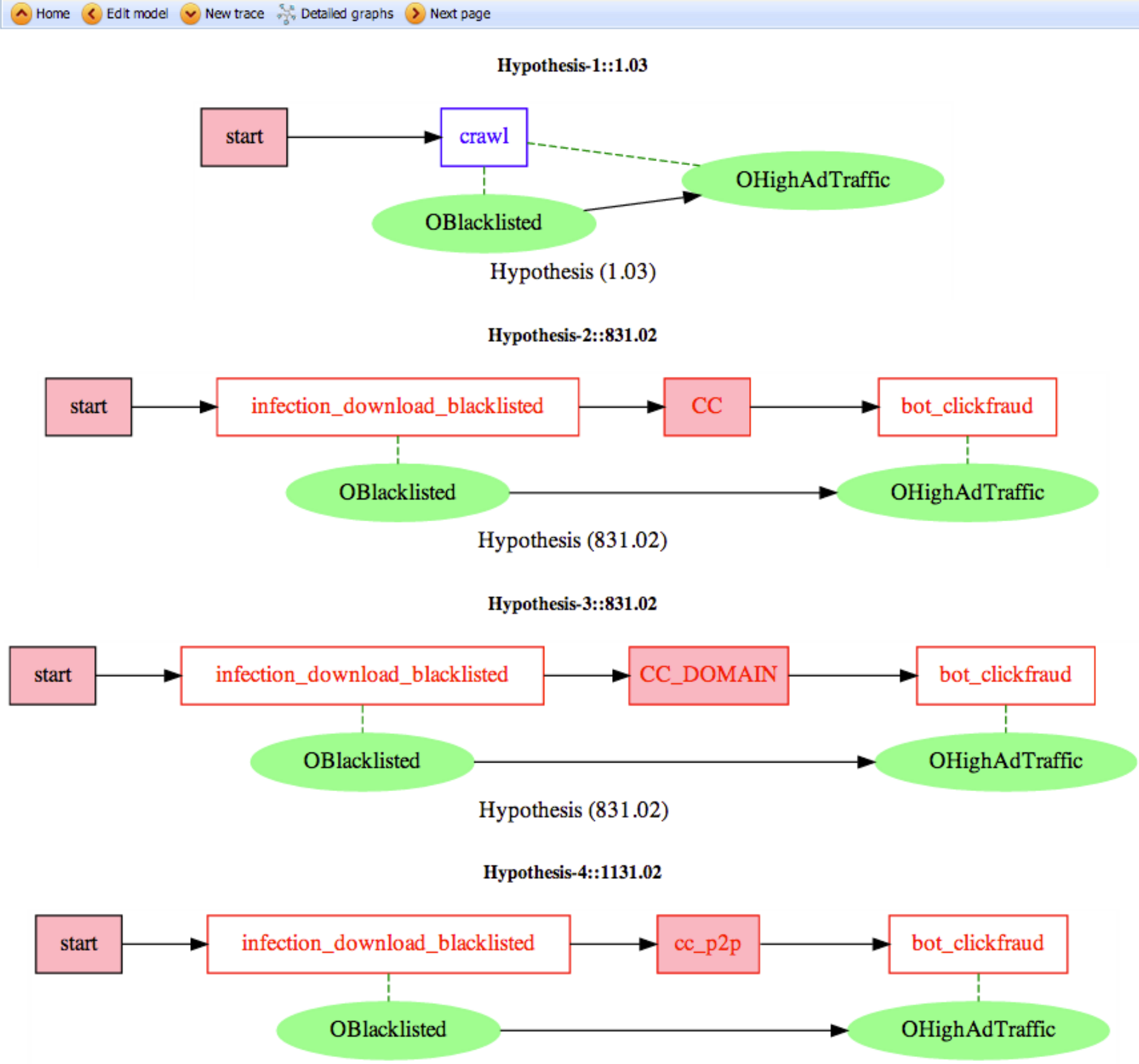}}  &
\raisebox{+0.1cm}{\includegraphics[width=0.8\columnwidth]{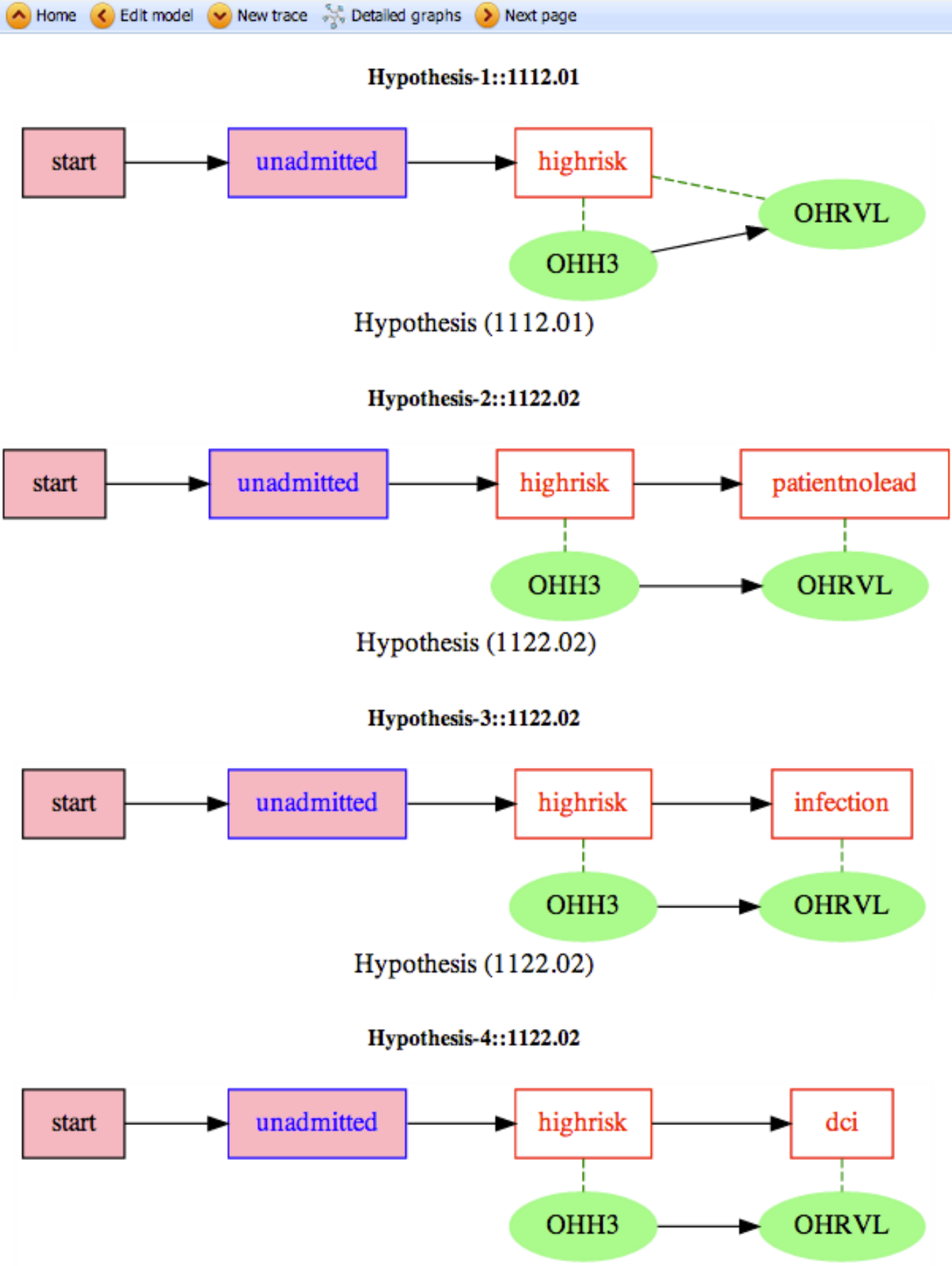}} \\
Figure 7 (c): Malware example 2 & Figure 7 (d): Intensive care example 1  \\
\hline
\end{tabular}

\label{fig:cyberexample}
\vspace{-0.49cm}
\end{center}
}
\end{figure*}

\setlength{\tabcolsep}{1pt} 
\begin{figure*}[ht!]
{\small
\begin{center}
\begin{tabular}{|c|c|}
\hline

\raisebox{+0.0cm}{\includegraphics[width=1.15\columnwidth]{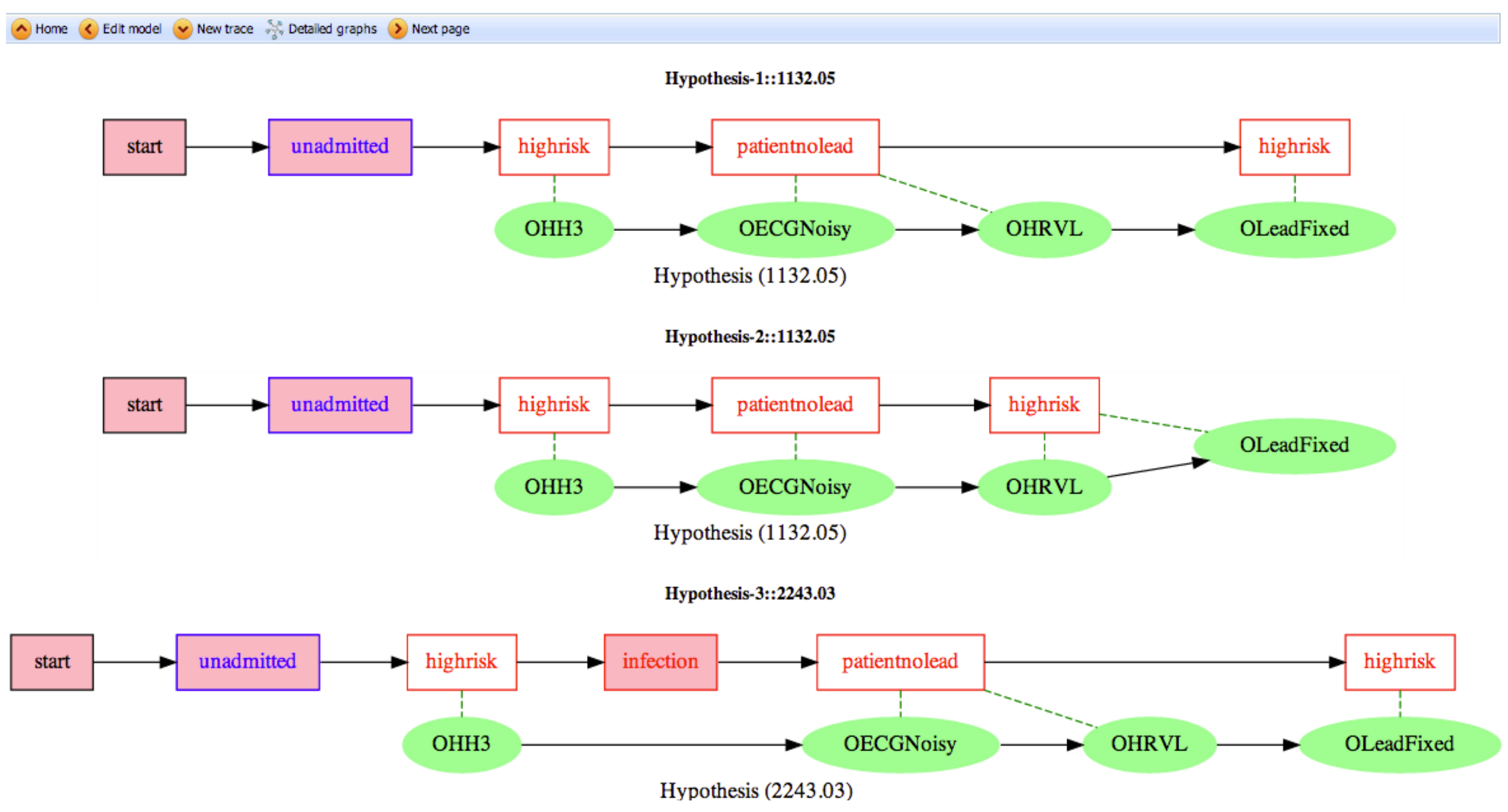}} &
\raisebox{+0.05cm}{\includegraphics[width=0.95\columnwidth]{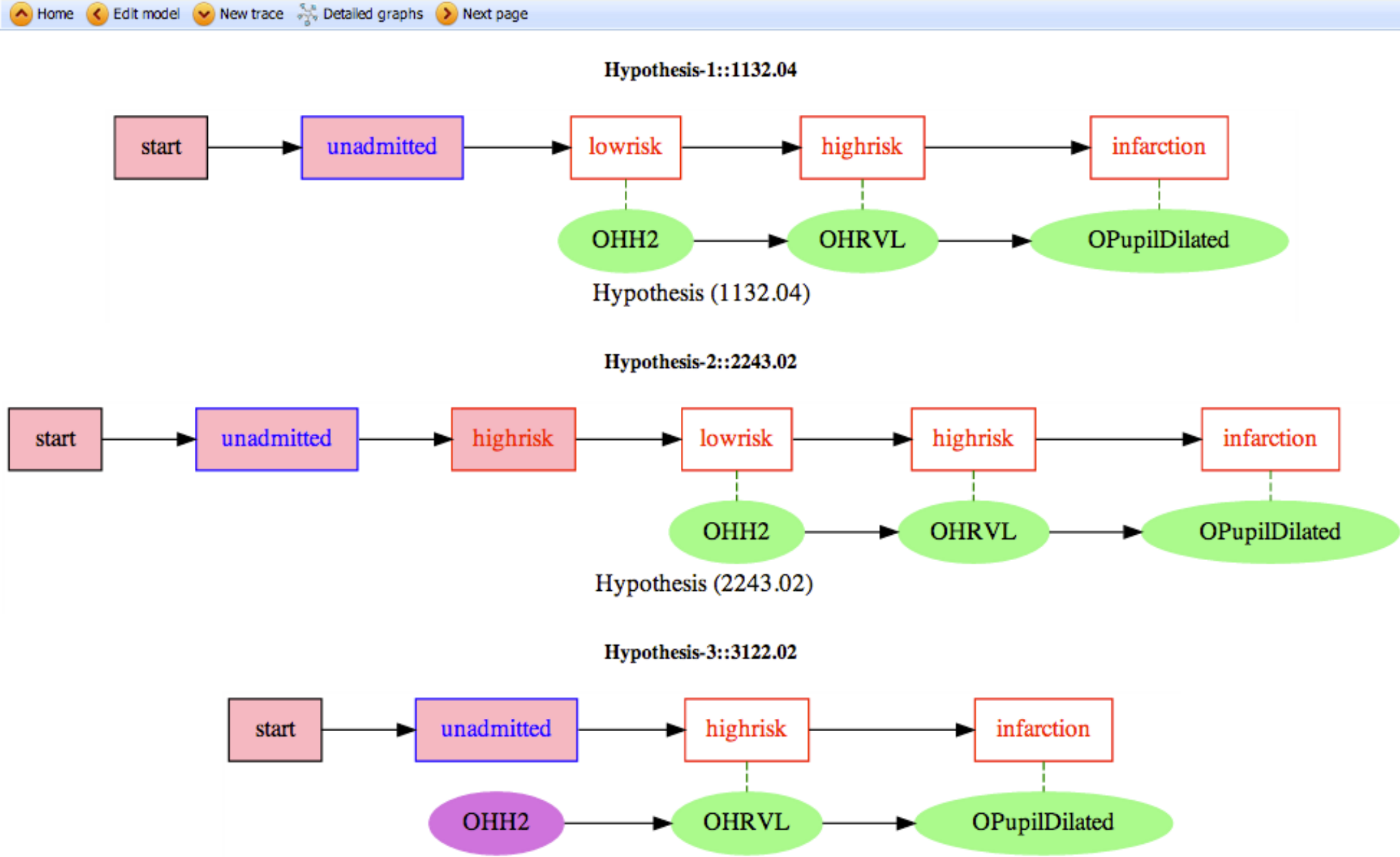}}  \\  
Figure 7 (e): Intensive care example 2 & Figure 7 (f): Intensive care example 3  \\

\hline
\end{tabular}

\label{fig:cyberexample}
\end{center}
}
\end{figure*}

\label{ide}
\subsection{LTS++ IDE}

LTS++ IDE is a web-based tool that helps users create planning problems by describing LTS++ models and generate hypotheses. LTS++ IDE consists of an LTS++ editor, graphical view of the transition system, specification of the trace, and generation of hypotheses. The tool automatically generates planning problems from the LTS++ specification and entered trace. The generated hypotheses are the result of running a planner and presenting the result from top-most plausible hypothesis to the least plausible hypothesis. 

Figure \ref{fig:ide} shows the LTS++ IDE. The top part is the LTS++ language editor which allows syntax highlighting and the bottom part is the automatically generated transition graph. The transition graph can be very useful for debugging purposes. LTS++ IDE also features error detection with respect to the LTS++ syntax. The errors and warning signs are shown below the text editor. They too can be used for debugging the model creation as part of step 8.

Figure \ref{fig:cybermodel} and \ref{fig:healthcaremodel} shows the LTS++ model for the malware detection and intensive care applications from Figure 1 respectively. The states are shown in blue with hyperstates specified in all caps. The observations are specified within the curly brackets and are shown in green. You can specify multiple observations by using space or comma between observations (see line 6). The state types are specified within angle brackets (see line 2). The transitions between states are specified using arrows. 
Each transition needs to be specified within a hyperstate.
Multiple transitions between states within a hyperstate can be specified using the vertical bar. The default state type is specified in line 1 and the starting state is specified in the last line.


\label{result}
\section{Generating Hypotheses via LTS++ IDE}

In this section, we will first explain how observations can be entered into the LTS++ IDE and then
we will go through a number of examples for both of our applications, and explain how to interpret the generated
results. This is the final step of the LTS++ model creation (i.e., step 9, testing).

Observations can be entered by clicking on the ``Next: edit trace" from the LTS++ IDE main page shown in Figure \ref{fig:ide}. Figure 7 (a) shows an example where the first observation is selected to be a download of an executable, and the second observation is now being selected from the drop-down menu.
Once the trace selection is complete, the hypotheses can be generated by clicking on ``Generate hypotheses".
The hypotheses are presented to the user 10 per page, and users can navigate through these pages. The next 10 hypotheses are generated once the user clicks on the ``Next page". 
 Note, the trace editor is intended mainly for testing purposes, and in operation the system will read observations automatically from an input queue. 

Figure 7 (b-f) show sample example runs for the malware and intensive care examples; these results are automatically generated by our tool. Each hypothesis is shown as a sequence of states matched to observed event sequence (via green dashed lines). The observations that are explained by a state are shown in green ovals, and unexplained observations are shown in purple. The arrows between the observations show the sequence of observations in the trace. The states shown in red are the bad states and good states are drawn in blue. Each hypothesis is associated with a cost. The lower the cost, the more plausible is the hypothesis.

Figure 7 (b) shows the top 3 generated hypotheses for the trace selected in Figure 7 (a). 
Our first hypothesis explained both observations. The second hypothesis, almost as plausible, shows infection followed
by the CC state. The third hypothesis leaves the second observation unexplained. 
In some instances, hypotheses include states that are not linked to any observation. For example, the \emph{CC}, \emph{CC\_Domain}, \emph{cc\_p2p} are the unobserved states in the non-crawling hypotheses in Figure 7 (c). 
Figure 7 (d) shows the automated generated results (the top-4) for an ambiguous observation HRVL.
The result of more specific, less ambiguous observation traces are shown in Figure 7 (e,f).

\setcounter{figure}{10}

\section{Summary and Discussion}

In this paper, we address the knowledge engineering problem of hypothesis generation motivated by two applications: malware detection and intensive care delivery. To this end, we proposed a modeling language called LTS++ and a web-based tool that enables the specification of a model using the LTS++ language. We also proposed a 9-step process that helps provide guidance to the domain expert in specifying the LTS++ model. Our tool, LTS++ IDE, features syntax highlighting, error detection, and visualization of the state transition graph. The hypotheses are generated by running a planner capable of generating multiple high-quality plans for the translated LTS++ model and the provided trace. The hypotheses can be visualized and shown to the analyst (doctor or network administrator), or can be further investigated automatically via the \emph{Strategic Planner} (see the Architecture Section) to run testing or preventive actions. 

In terms of evaluation of our model, we have worked with users outside of our group to develop different LTS++ models in different domains. The feedback we received from them is positive and helped us improve our tool and the creation process. Particularly, one of models developed this way is now used within a larger application.

Our approach in using planning is related to several approaches in the diagnosis literature in which the use of planners as well as SAT solvers is explored (e.g., \cite{GrastienARKAAAI07,sohBiaMcIKR10}). 
In particular, the work on applying planning for the intelligent alarm processing application is most relevant \cite{GrastienDX211,Haslum-Two-11}. The authors have considered the case where they can encounter unexplainable observations, but have not provided a formal description of what these unexplainable observations represent or how the planning framework can model them. In this work we address this, as well provide tools for domain experts and introduce a simple language that can be used instead of PDDL.

\bibliographystyle{aaai}

\begin{small}
 \providecommand{\Proceedings}{Proceedings }
  \providecommand{\International}{International }
  \providecommand{\longshort}[2]{#1 (#2)}
  \providecommand{\longshortnopar}[2]{#1}

\end{small}

\end{document}